# SCAFFOLD: Toward Warrant-Preserving Evidence Transport for Reusable LLM-Agent Skills

Nami Cho*
*Columbia University (alum), USA*
nc3009@columbia.edu

Shailja Gupta*
*Carnegie Mellon University (alum), USA*
shailjag@tepper.cmu.edu

***Abstract*—An agent skill that succeeded yesterday may be executable today without being justified for reuse today. Tests may cover only observable contract clauses, outcome oracles may be absent, evaluations may inherit the same erroneous specification, and permissions or dependencies may change between validation and invocation. We argue that the appropriate unit of prospective trust is a Warranted Reuse Claim (WRC): a defeasible, claim-specific authorization tied to a versioned skill, target context, evidence, assumptions, and identified defeaters. The paper sharpens this position through Warrant-Preserving Evidence Transport (WPET), a proposed discipline for deciding which historical evidence remains admissible after context change. It distinguishes evidence admissibility from truth of the supported safety or effectiveness claim; states a conditional preservation result and its limits; specifies tri-valued applicability and asymmetric context-transfer judgments; and illustrates how provenance dependence and policy-invalidating changes alter authorization. We explicitly relate the proposal to assurance cases, contracts, and skill-library maintenance rather than claiming novelty for their ingredients. A falsifiable longitudinal evaluation would compare WRC-guided policies with strong version-aware baselines at matched coverage, utility, and review cost. This paper contributes a formalized problem statement, testable design obligations, and reporting protocol—not an implemented verifier, empirical superiority result, or proof of universal agent safety.**

## I. Introduction

Reusable skills—scripts, procedures, tool-use recipes, workflow graphs, and instructions—make autonomous agents more efficient by externalizing successful behavior for later invocation. Their persistence also creates a temporal mismatch: evidence is acquired in a past context while authorization is demanded in a new one. A previously passing data-cleaning routine can silently delete newly valid anonymous-customer records; a backup artifact can satisfy file-format checks while being unrestorable; a refund workflow can receive an API success response despite violating a newly applicable approval rule. The problem is not merely whether a skill runs, but whether evidence about an earlier execution is sufficiently relevant to justify a particular future use.

We advance a bounded thesis: an authorization decision should be explainable as a prospective claim supported by provenance-aware evidence under disclosed context-preservation assumptions, and should be retractable when those assumptions fail. We do not claim that this thesis is unprecedented in assurance research, that it requires an autonomous agent-specific proof logic, or that more governance is always beneficial. The scientific issue is whether claim-specific evidence transport can improve the harm–utility–cost frontier relative to mature contract and maintenance approaches.

Contributions and boundaries. (C1) We define a prospective evidence-transport problem for versioned agent skills and distinguish compliance, goal success, and authorization. (C2) We propose the WRC and WPET representations, including explicit preservation obligations, unknown applicability, provenance dependence, and defeaters. (C3) We provide a conditional evidence-admissibility proposition and counterexamples that clarify what it does not guarantee. (C4) We specify a discriminating, falsifiable evaluation against strong contemporary baselines. These are conceptual and methodological contributions; no experiments were conducted.

## II. Related work

Design by contract and runtime monitors verify properties stated in specifications; they cannot on their own establish that a partial specification captures the user's underlying goal. The oracle problem and metamorphic testing literature explain why passing checks can coexist with incorrect real-world outcomes [1], [2]. Provenance models, policy-based access control, change-impact analysis, and assurance cases likewise long predate this paper. Assurance 2.0-related research already analyzes assurance claims, evidence, and their semantic coherence, including doubts and defeaters [3]; we therefore do not claim to have invented defeasible claim-based assurance.

Incremental assurance explicitly studies how supporting evidence and residual doubts evolve over time [4]. Safe regression-test selection likewise studies when a subset of earlier tests can be rerun after software changes instead of executing the entire test suite [5]. WPET therefore does not introduce selective revalidation as a general idea. Its proposed research question is whether claim-specific, defeater-aware evidence transport provides a useful and reproducible authorization discipline for evolving LLM-agent skills beyond those established approaches.

*Both authors contributed equally.**

Agent-skill literature now provides particularly important comparators. SkillOps models typed skill contracts and dependency-aware library maintenance [6]. Metere presents verification levels and capability-gating concepts [7]; Contractual Skills emphasizes inspectable enterprise requirements [8]; and the Dynamic Agent Skills survey synthesizes verification, admission, provenance, maintenance, and rollback [9]. Voyager demonstrates durable skill libraries in an embodied-agent setting [10]. These works limit any claim of priority over contracting, verification, and lifecycle governance. Our candidate differentiator is narrower: an explicit, claim-specific account of when an old evidence item remains admissible in a materially changed target invocation, with observable grounds for refusing transport.

TABLE I. Existing capabilities and open comparisons

| Existing capability | What it supplies | Open comparison for SCAFFOLD |
|---|---|---|
| Strong contracts and policy gates | Version-aware clauses, permissions, approvals | Does explicit evidence transport add better decisions or explanations at equal cost? |
| Assurance cases / defeaters | Claims, arguments, evidence, doubts | Can change-sensitive skill invocation be checked with less manual effort? |
| SkillOps / skill maintenance | Typed contracts and library-level evolution | Does claim-specific transport outperform maintenance triggers? |
| Outcome testing / independent oracles | Grounded evidence of task completion | Which historical outcome evidence remains relevant after drift? |

The novelty test is deliberately stringent: if an incremental-assurance, selective regression-testing, or version-aware contract baseline instantiates the same preservation and defeater judgments at comparable effort, WRC may be useful terminology or reporting infrastructure rather than a new decision capability. That is a valid negative result, not an exception to be dismissed.

## III. Problem Formulation: Evidence Governed Skill Reuse

Let s denote an immutable, version-identified skill artifact; c a target invocation context; φ a prospective proposition such as "s may execute under the specified privileges and side-effect limit"; E a time-stamped evidence ledger; A declared applicability assumptions; and D known or discoverable defeaters. A Warranted Reuse Claim is the record WRC = ⟨s, c, φ, E, A, D⟩. It is not a scalar confidence score and is not equivalent to a statement that φ is true in every unobserved execution.

For scope-sensitive comparison, represent the target context as $c = (g, X, m, t, p, b, v, \ell, q)$, including goal g, input family X, model/runtime m, tools t, privileges p, side-effect boundary b, dependency and policy versions v, consequence profile ℓ, and time or environment state q. This schema is intentionally extensible: omitted context dimensions can invalidate claims.

For an execution trace τ, let O(τ) be the portion observable to monitors, let C(τ) indicate satisfaction of monitored clauses, and let G(τ) indicate independent goal achievement. In general:

$$C(\tau) = 1 \not\Rightarrow G(\tau) = 1. \tag{1}$$

A monitored pass supports only the checked predicates and any conclusions licensed by separately defended assumptions. If the goal depends on an unobserved state, or if the contract omits a goal-relevant constraint, monitor-indistinguishable traces can differ in real outcome. This is the central specification-adequacy problem, not an indictment of correctly designed monitors.

## IV. Warrant-Preserving Evidence Transport (WPET)

### A. Evidence Representation and Admissibility

An evidence item e records its claim tested, skill and contract versions, source or evaluator, test context $c_e$, result (pass, fail, inconclusive), environmental fingerprint, detectable failure modes, timestamp, and known limitations. Let Support(e, φ | A) assert that the methodology and declared assumptions make e relevant to φ in its original context. Support is not synonymous with evidence being sound, independent, or decisive.

Define $\mathrm{App}\square(e, s, c, \varphi \mid A) \in \{1, 0, ?\}$, where 1 denotes admissible support under policy P and justified transport, 0 denotes disconfirmed or inapplicable support, and ? denotes insufficient information. An unresolved value is not positive evidence. The usable support set is:

$$E^{+}_{p}(s, c, \varphi \mid A) = \{\, e \in E : \mathrm{App}_{p}(e, s, c, \varphi \mid A) = 1 \,\}. \tag{2}$$

Importantly, a failing test is not discarded when it is "inapplicable" as positive support: relevant negative findings are separately evaluated as possible defeaters. Otherwise a policy could hide adverse evidence simply by filtering it out of $E^{+}$.

### B. Contextual Preservation Obligations

The substantive question is what must remain invariant for a particular evidence-to-claim inference. Let R(e,φ) be a finite set of context-relevant premises on which this inference depends. Examples include matching authorization policy, stable identity and semantics of an external API, a valid outcome oracle, or unchanged side-effect boundaries. Let $\Delta(c_1,c_2)$ be the documented change set. Define $\mathrm{Pres}\square(r,c_1,c_2) \in$ {true,false,unknown} as a policy-approved check of whether premise r remains applicable in $c_2$. Unknown changes or unobservable dependencies force an unresolved result rather than optimistic preservation.

$$\begin{aligned} &\mathrm{WPET}_{p}(e, c_1 \rightarrow c_2, \varphi \mid A) = 1 \ \text{ iff} \\ &\mathrm{Support}(e, \varphi \mid A) \text{ and } \forall r \in R(e, \varphi): \\ &\mathrm{Pres}_{p}(r, c_1, c_2) = \text{true}. \end{aligned} \tag{3}$$

This is a proposed admission rule, not a novel mathematical identity. Its implementation must make each r and its check concrete. In particular, a developer cannot simply define R(e,φ) to omit inconvenient dependencies. A soundness claim would require a separately justified requirement-completeness assumption; without it, WPET may transport evidence under a false premise.

The directed claim-specific transport judgment $T\square(c_1 \rightarrow c_2, \varphi \mid A) \in$ {valid,invalid,unresolved} summarizes item-level preservation obligations for a selected evidence family. Transport need not be symmetric: a read-only evaluation does not license write privileges, and evidence about a production policy does not necessarily certify behavior under a less privileged test stub.

### C. Conditional Evidence Preservation

Proposition 1 (conditional evidence-admissibility preservation). Fix claim $\varphi$, evidence e acquired in $c_1$, and an evidence-to-claim inference rule I. Assume: (i) I validly licenses e as admissible support for $\varphi$ in $c_1$; (ii) $R(e,\varphi)$ contains every premise whose context-specific truth is required for I; (iii) each premise in $R(e,\varphi)$ is preserved in $c_2$; (iv) skill and evidence identities are intact; and (v) no applicable defeating observation invalidates e or I. Then e remains admissible as support for $\varphi$ under the same inference rule in $c_2$.

Proof sketch. By (ii), all context-sensitive premises needed for I are represented in R. By (iii) they remain satisfied in $c_2$; by (iv) the object and evidence referenced by I are unchanged; by (v) no applicable defeater cancels the inference. Therefore the premise conditions required to invoke I still hold in $c_2$, and the admissibility conclusion follows. □

Interpretation and limit. This is a conditional preservation argument, not a new safety theorem: the strong completeness assumption (ii) carries most of the burden. The proposition neither establishes that $\varphi$ is true nor guarantees that identifying R is decidable or cheap. The technical research opportunity is a useful, conservative, checkable approximation to R and Pres that survives strong baseline comparison. A reviewer should reject any assertion of broad formal safety based only on this proposition.

### D. Composition and Failure Modes of Evidence Transport

Suppose evidence passes from $c_1$ to $c_2$ and then to $c_3$. Pairwise transport is not automatically compositional if a latent dependency was never represented, or if preservation checks are relative to observations that become stale. A sufficient compositional condition is that the full set $R(e,\varphi)$ remains stable, premises are checked against the same semantic interpretation, and preservation holds for both transitions without unrecorded defeaters. Under these assumptions, the item is admissible in $c_3$; without them, transport must be independently reassessed. A practical implementation should store the dependency and change ledger that makes these conditions inspectable.

Counterexample A (hidden dependency). A backup test validates file existence and checksum but omits restoration capability from R. A database serialization upgrade preserves both recorded premises yet breaks restore. WPET returns an incorrect positive verdict because the premise set is incomplete. This illustrates the method's fundamental limitation, not an unexpected edge case.

Counterexample B (asymmetric privileges). A read-only report skill validated on staging is invoked with production delete permissions. Even if the skill text is unchanged, the enlarged side-effect boundary invalidates a previously sufficient confinement premise. Old schema evidence may transport, whereas authorization evidence must not.

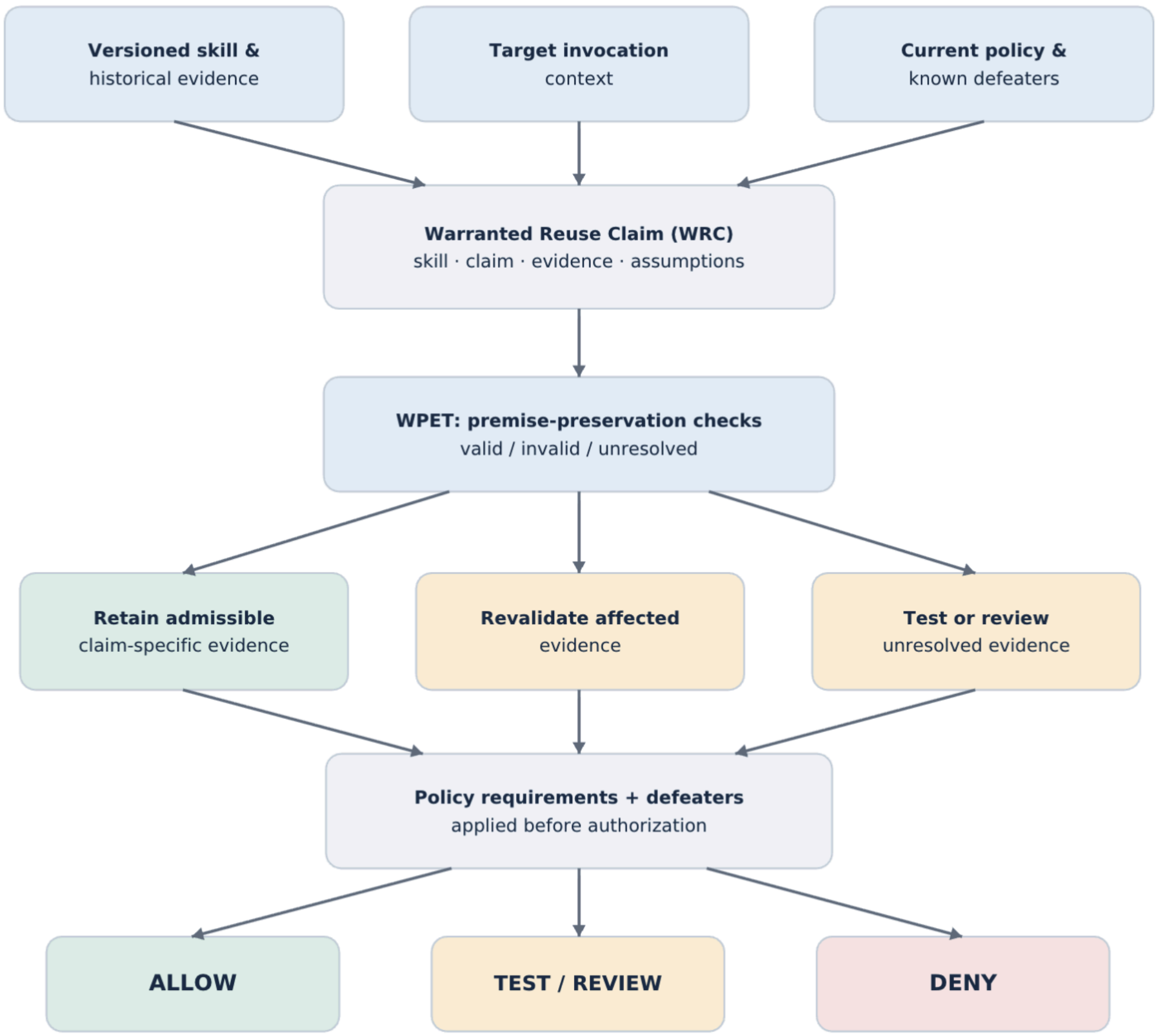

Fig. 1. Claim-specific evidence transport and reuse authorization. Illustration of the proposed WRC / WPET process (Sections III–V). Evidence transport does not establish claim truth or guarantee safe execution. Allow depends on all policy requirements being satisfied, no active hard prohibitions, and no unacceptable unresolved uncertainty.

## V. Defeasible Authorization and Evidence Dependence

### A. Provenance and Dependence Among Evidence Sources

Evidence can share a common originating error. If one model writes a skill, infers its specification, generates examples, and judges the examples, multiple favorable assessments may all inherit the same false goal interpretation. Let $e_i \sim e$□ denote documented provenance or shared-failure-mode dependence; the graph records why two judgments may not be independent. It does not prove statistical independence for disconnected nodes. For exchangeable variables with common pairwise correlation ρ, the illustrative variance-equivalent sample size is $n/[1+(n-1)\rho]$; this familiar formula is not a proposed WRC confidence estimator and is not directly applicable when evaluator correlation is unknown.

An implementable policy should disclose evaluator identity, source data, oracle origin, and detectable failure modes; require independent environment outcomes where proportionate; and treat absent or inconclusive tests as uncertainty. Diversity of provenance is a hypothesis to test, not evidence of correctness by fiat.

### B. Defeasible Authorization Semantics

Let Req□(φ,c) list the evidence and boundary requirements for a target claim, Sat□($E^+$,Req□) indicate their satisfaction, Hard□(D,c,E) identify triggered prohibitions, and Unresolved□(c,E) identify uncertainties unacceptable for direct execution. Define a three-way policy:

$$\delta_p(s, c, E, A, D) = \begin{cases} \text{deny} & \text{if } \mathrm{Hard}_p \neq \varnothing; \\ \text{allow} & \text{if } \mathrm{Hard}_p = \varnothing \wedge \mathrm{Sat}_p(E^+, \mathrm{Req}_p) \\ & \wedge \neg \mathrm{Unresolved}_p; \\ \text{test/review} & \text{otherwise.} \end{cases} \tag{4}$$

Hard prohibitions dominate positive tests. The policy should distinguish temporary refusal from permanent retirement, and should identify who can resolve an uncertainty. A newly discovered defeater requires a reassessment, but not every dependency update necessitates denial. If a mandatory approval is missing, the correct outcome may be “review required before execution,” becoming denial only if approval cannot be obtained under the policy.

Proposition 2 (scope fidelity by enforcement, conditional). If an allow decision binds an immutable skill version and an enforced capability boundary B, and all executable tool calls are mediated by that boundary, the authorized execution cannot invoke capabilities outside B. This is a property of the external enforcement mechanism under complete mediation, not a consequence of evidence-ledger notation. SCAFFOLD therefore complements least privilege and sandboxing rather than replacing them.

Proposition 3 (defeater responsiveness by policy). If Hard□ is recomputed before every authorization decision from an up-to-date authoritative source and a newly discovered critical defeater is reflected in that source, Equation (4) prevents a subsequent allow decision while that defeater remains active. A time-of-check/time-of-use race or stale source breaks the premise. This is a straightforward property of the rule; implementation and race handling remain open.

## VI. Illustrative Cases and Selective Evidence Revalidation

Consider skills validated for refunds no greater than \$100 under regional policy v1, without supervisor approval. Its tests establish API schema validity, correct order lookup, and compliance with the v1 authorization clause. A future invocation requests a \$150 refund in another region with policy v2 requiring supervisor approval. A weak contract-only gate checking the unchanged executable and old tests allows execution. A strong version-aware gate may correctly block the request—this is why the example is not evidence of WRC superiority.

TABLE II. Claim-specific evidence after a refund policy change

| Claim / evidence | Source context | Transport judgment | Decision implication |
|---|---|---|---|
| API payload schema check | s, API schema unchanged | Valid if API semantics unchanged | Retain for schema claim only |
| Refund ≤ \$100 tested | v1, ≤ \$100 | Invalid for \$150 amount | Fresh amount-policy check |
| Approval not required | Region A, v1 | Invalid under Region B, v2 | Supervisor approval required |
| External settlement outcome | Earlier settled refund | Unresolved for different jurisdiction | Fresh outcome validation if necessary |
| Shared-model test suite | One model-created specification | Dependent; not independent confirmation | Disclose provenance / seek external check |

A complete WRC records the exact authorization claim, the retained schema evidence, the rejected policy evidence, the unsettled outcome claim, and the active approval defeater. The appropriate decision is review before execution or denial when mandated approval cannot be obtained. The discriminating comparison is not whether a sophisticated policy system notices the approval rule; it is whether explicit claim-level transport improves auditability, selective revalidation, risk outcomes, or engineering cost.

### A. Policy-Driven Evidence Invalidation

The refund example above demonstrates claim-level invalidation after a jurisdictional policy change. It does not establish that WPET is superior to a strong policy-aware gate: both could demand supervisor approval. The following constructed case instead isolates the potential benefit of retaining unaffected evidence.

### B. Selective Evidence Transport under Dependency Changes

Constructed scenario (not empirical evidence). Consider an immutable invoice-reconciliation agent skills with 12 previously satisfied verification obligations, divided into four claim families with three independent checks each: F1, identity and privilege confinement; F2, source-document extraction fidelity; F3, tax-calculation correctness; and F4, immutable audit-log delivery. A new tax-engine library version changes rounding semantics while the skill, privileges, extraction service, and audit-log service remain fixed. The target context $c_2$ differs from $c_1$ only in the tax-engine dependency and its documented semantics. Under an explicit, reviewed dependency map, only F3 has a premise requiring the old rounding behavior.

For each evidence item $e_i$, let $R(e_i, \varphi_i)$ enumerate the premises necessary for that item to support its specific claim. WPET evaluates those obligations against the change $\Delta(c_1,c_2)$. Assume independent change records, trustworthy service fingerprints, and a complete dependency map for these 12 obligations. Under these assumptions, the nine F1/F2/F4 results remain admissible as evidence of their narrow claims; the three F3 results become invalid or unresolved and require fresh tax-engine validation. Historical acceptance of F1/F2/F4 does not by itself prove the combined invoice outcome in $c_2$; an integration or end-to-end check may still be mandated by the target policy.

TABLE III. Comparison under stated assumptions

| Strategy | Retesting decision | What the case demonstrates |
|---|---|---|
| Full context-level revalidation | Repeat all 12 checks | Safe conservative reference under the scenario assumptions |
| Version-triggered blanket revalidation | Repeat all 12 checks on any relevant dependency change | May incur avoidable work when the trigger is coarse |
| WPET with complete claim-dependency mapping | Retain 9 narrowly applicable results; rerun 3 tax checks, plus any required integration check | Potential selective-revalidation benefit; not a superiority result |
| Strong dependency-aware contract / assurance baseline | Can also retain 9 and rerun 3, if equipped with the same mapping and rules | Equal performance is entirely possible and must be measured |

Illustrative cost accounting. If each obligation costs $k$ to recheck, full or blanket retesting costs $12k$, while selective retesting costs $3k + h$, where $h$ is the cost of obtaining and checking the dependency map, transport premises, provenance, and any additional mandated integration assessment. The illustrative net saving is $9k - h$, positive only if $h < 9k$. This arithmetic deliberately excludes any shared verification overhead already present in both strategies; such costs must be included consistently in an actual comparison. If all strategies require a common end-to-end test with cost $j$, add $j$ to both totals; it cancels only when it is genuinely identical across strategies.

A more general characterization is: let $V$ be the set of historical verification obligations, $I(\Delta,\varphi) \subseteq V$ the obligations whose claim-relevant premises are changed, defeated, or unresolved, and $Cost(Q)$ the cost of rechecking a subset $Q$. A conservative selective policy retests $I$ and preserves evidence outside $I$ only when all preservation checks return true. Its claimed efficiency condition is $Cost(V) > Cost(I) + h$, with matched coverage obligations. This is a conditional engineering hypothesis, not a theorem of superior safety or an empirically established inequality.

Comparison to prior work. The selective-revalidation idea itself has substantial precedent in safe regression-test selection [5] and incremental assurance [4]. The proposed contribution here is the explicit transport of claim-specific evidence, including provenance, contextual permissions, and potential defeaters; a strong baseline with equivalent information may make identical decisions. Any efficiency claim must therefore measure the incremental benefit of these records and judgments rather than attributing all avoided tests to WPET.

Failure conditions and discriminating prediction. If the rounding change also influences extraction normalization or audit formatting through an undocumented dependency, the nine retained results may not be applicable; the claimed saving could be unsound. If dependency review costs exceed the avoided tests, or if strong incremental-assurance and regression-test-selection methods [4], [5] already make the same selective decision, WPET offers no demonstrated decision advantage. Its potential distinctiveness then rests on standardized claim-level evidence-transport records and independent reproducibility. A study should randomize localized versus broad changes, measure revalidation cost and missed-impact defects, and compare all methods using identical dependency information and independent outcome adjudication.

Interpretive boundary. The example compares the amount of evidence revalidated, not the truth of the resulting authorization claim. Authorization remains conditional on policy-required outcome checks, current prohibitions, and independently verified execution boundaries. This construction does not report observations from deployed agents.

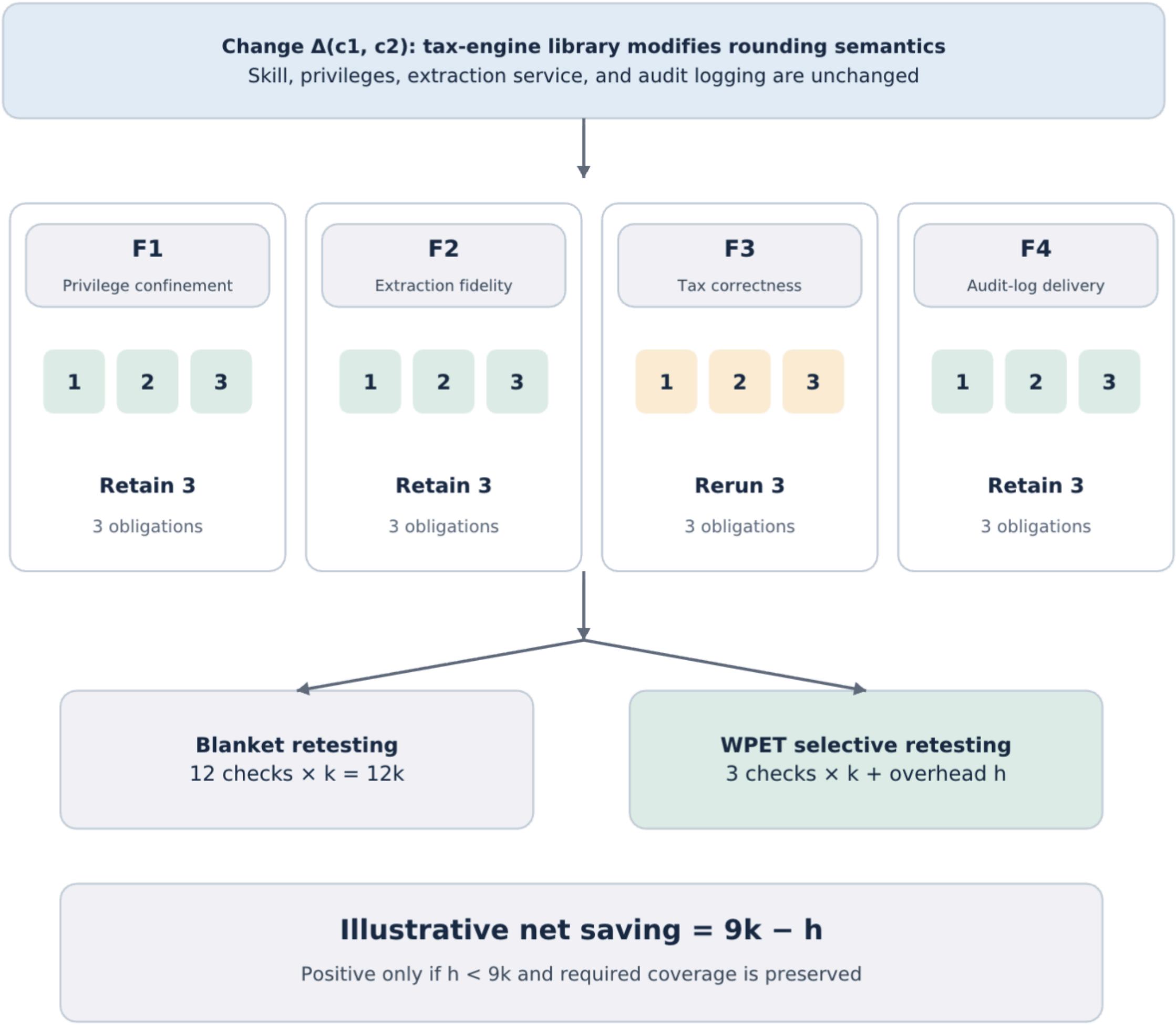

Fig. 2. Selective evidence transport after localized tax-engine change. Constructed invoice-reconciliation scenario (Section VI-B). This illustration does not report measured results. Selective reuse requires an accurate claim-dependency map, trustworthy change records, unchanged relevant premises, and current defeater checks. A common end-to-end integration check may still be required. Strong dependency-aware baselines can make the same selection; no superiority is demonstrated.

## VII. Falsifiable Hypotheses and Evaluation Methodology

We propose a longitudinal evaluation in which agents acquire reusable skills on an initial task stream, development thresholds are frozen, and independent investigators present held-out invocations with schema drift, policy changes, dependency updates, stale data, incomplete specifications, hidden side effects, and partially observed environmental goals. The unit of measurement is an eligible reuse opportunity, not a test invocation selected after results are known. Outcome adjudication must be independent of the method authoring the skill and its tests.

Baselines should include (B0) unrestricted reuse, (B1) model self-assessment, (B2) a strong version-aware contract and approval gate, (B3) outcome-only gating where available, (B4) a skill-library maintenance approach such as SkillOps, and (B5) an assurance-case-inspired policy instantiated with equivalent contextual information [3], [4]. A strong change-impact/regression-test-selection comparator should also be implemented with matched dependency information [5]. All approaches should have matched models, opportunity streams, access to external evidence, and compute or human-review budgets. A budget-limited configuration of WRC should be tested, not only an unconstrained one.

TABLE IV. Hypotheses and disconfirming results

| Hypothesis | Primary evidence for | Disconfirming result |
|---|---|---|
| H1: selective transport reduces redundant revalidation | Fewer tests at equal harmful reuse and useful completion | Strong contract baseline equals or beats cost/quality |
| H2: provenance-aware evidence improves decisions | Better held-out failure detection at fixed review cost | Provenance contributes no incremental signal |
| H3: defeater tracking reduces harmful drift | Lower harm at matched completion and cost | Simple version triggers perform equally well |
| H4: explicit warrants improve audit reconstruction | Independent reviewers reproduce decisions more accurately | Baseline audit trails are equally interpretable |

Let $\pi$ map opportunities to allow, test, review, or reject; $H_\pi$ be adjudicated harm, $U_\pi$ useful completions, and $K_\pi$ evaluation/review burden per eligible opportunity. A suitable benchmark is:

$$\begin{aligned} &\text{minimize}_\pi \;\; E[H_\pi] \\ &\text{subject to} \;\; E[U_\pi] \geq u_{min} \;\; \text{and} \;\; E[K_\pi] \leq k_{max}. \end{aligned} \tag{5}$$

This constrained objective is established decision-analysis practice, not a novel optimization algorithm. Report harm–utility–cost frontiers, false denials, total latency, computational cost, human effort, and uncertainty intervals clustered by skill and task family. Pre-register primary endpoints, consequential harm categories, review thresholds, and task-family splits. Report both intention-to-govern (all eligible opportunities) and executed-opportunity metrics; the latter alone rewards blanket refusal.

A useful ablation would remove (a) context-preservation obligations, (b) outcome evidence, (c) provenance dependence, and (d) active defeater checks separately. Another should introduce specification misspecification to measure whether explicit warrants merely formalize the wrong assumptions. WPET fails its main applied claim if a strong baseline matches or exceeds its risk–utility performance at equal cost, or if preservation obligations require impractical expert attention.

## VIII. WRC Reporting and Reproducibility Protocol

For each skill version and evaluation context, report: the prospective claim and authorized boundaries; the evidence items and evaluator provenance; the original and target context fingerprints; the purportedly preserved premises $R(e,\varphi)$; the verification method for each premise; the unknown or invalidated dependencies; any observed counterevidence and hard defeaters; the decision and reviewing actor; the independent outcome, harm and completion measurement; and the direct and human evaluation costs. Record timestamps and policy versions so that authorization can be reconstructed after the environment changes.

Studies should explicitly distinguish evidence admissibility, plausibility of the claim, enforced capability scope, and actual downstream outcome. An evidence ledger should never be presented as a safety certificate. In domains lacking a reliable oracle, the absence must remain visible; model self-confidence is not an equivalent replacement.

## IX. Limitations and Threats to Validity

First, premise identification may be incomplete: unknown dependencies, unobserved state, misleading logs, or external-policy ambiguity can make transport unsound. Second, WRC may impose overhead greater than the cost of re-executing the skill, especially for low-impact reversible tasks. Third, the semantics presuppose trustworthy registries of versions, policy sources, identity, and event provenance; attackers can target these registries. Fourth, correctness and legal or organizational authorization are separate goals, and neither is guaranteed by static compliance. Fifth, there is no universal objective outcome oracle for creative or strategic tasks. Sixth, the framework as written does not resolve time-of-check/time-of-use races or multi-agent delegation; claims should be rechecked at an appropriate enforcement boundary.

Finally, assurance cases, policy engines, and version-aware contracts may already realize much of this discipline. WPET should be adopted only where its explicit transport obligations improve an outcome that matters: selective validation cost, prevention of harmful reuse, decision reconstruction, or scope integrity. If not, its strongest remaining contribution is a common reporting format and sharply formulated research agenda.

## X. Conclusion

SCAFFOLD proposes that the unit of trust for persistent LLM-agent skills should be a Warranted Reuse Claim about one prospective invocation, not a permanent quality label earned by prior successes. Warrant-Preserving Evidence Transport refines that position by demanding explicit, claim-relevant preservation obligations before historical evidence is carried into a changed context. Its conditional preservation argument concerns admissibility of evidence, not safety of a future execution; incomplete premises and shared evaluator errors remain fundamental risks. A decisive next step is a matched, longitudinal comparison against strong contract, assurance-case, and lifecycle-maintenance baselines. Until such evidence exists, SCAFFOLD should be understood as a precise and falsifiable position rather than a validated algorithm or guarantee.